\documentclass{article}
\PassOptionsToPackage{numbers, compress}{natbib}

\usepackage[preprint]{neurips_2024}

\usepackage[utf8]{inputenc} % allow utf-8 input
\usepackage[T1]{fontenc}    % use 8-bit T1 fonts
\usepackage{hyperref}       % hyperlinks
\usepackage{url}            % simple URL typesetting
\usepackage{booktabs}       % professional-quality tables
\usepackage{amsfonts}       % blackboard math symbols
\usepackage{nicefrac}       % compact symbols for 1/2, etc.
\usepackage{microtype}      % microtypography
\usepackage{xcolor}         % colors

\usepackage{bm}
\usepackage{amsmath}
\usepackage{amsfonts}
\usepackage{amssymb}
\usepackage{mathtools}
\usepackage{mathdots}
\usepackage{physics}
\usepackage{comment}

\usepackage{amsthm}
\usepackage{amssymb}
\usepackage{enumerate}

\usepackage{listings}

\usepackage[resetlabels,labeled]{multibib}
\newcites{SI}{Additional References}

\title{Phase Transitions in the Output Distribution \\ of Large
Language Models}

\author{
 Julian Arnold\\
  Department of Physics\\
  University of Basel\\
  Klingelbergstrasse 82, 4056 Basel, Switzerland \\
  \texttt{julian.arnold@unibas.ch}
  \And
   Flemming Holtorf\\
   CSAIL\\
  Massachusetts Institute of Technology\\
  Cambridge, Massachusetts 02139, USA\\
  \texttt{holtorf@mit.edu}
  \And
   \And
   Frank Schäfer\\
   CSAIL\\
  Massachusetts Institute of Technology\\
  Cambridge, Massachusetts 02139, USA\\
  \texttt{franksch@mit.edu}
  \And
    Niels Lörch\\
  Department of Physics\\
  University of Basel\\
  Klingelbergstrasse 82, 4056 Basel, Switzerland \\
  \texttt{niels.loerch@unibas.ch} \\
  }

\begin{document}

\maketitle

\begin{abstract}
In a physical system, changing parameters such as temperature can induce a phase transition: an abrupt change from one state of matter to another. Analogous phenomena have recently been observed in large language models. Typically, the task of identifying phase transitions requires human analysis and some prior understanding of the system to narrow down which low-dimensional properties to monitor and analyze. Statistical methods for the automated detection of phase transitions from data have recently been proposed within the physics community. These methods are largely system agnostic and, as shown here, can be adapted to study the behavior of large language models. In particular, we quantify distributional changes in the generated output via statistical distances, which can be efficiently estimated with access to the probability distribution over next-tokens. This versatile approach is capable of discovering new phases of behavior and unexplored transitions -- an ability that is particularly exciting in light of the rapid development of language models and their emergent capabilities.
\end{abstract}

\section{Introduction}
Colloquially, the term \textit{phase transition} refers to a change among the basic phases of matter. For example, in response to changes in external conditions such as temperature or pressure, water can transition to a solid, liquid, or gaseous state. More broadly, in physics a phase transition refers to an abrupt change in the macroscopic behavior of a large-scale system of interacting constituents~\cite{saitta:2011,sethna:2021}. Notable examples include transitions in the magnetic properties of materials~\cite{onsager:1944}, transitions from a normal conducting state to a superconductor~\cite{taillefer:2010}, transitions in the entanglement properties of quantum circuit~\cite{lavasani:2021}, or the collective motion of active matter such as a flock of birds~\cite{viscek:1995}. 

In the context of artificial intelligence, ``phase transition''-like phenomena have also been observed in the learning behavior of neural networks (NNs)~\cite{saitta:2011,shwartz:2017,gur:2018,mcgrath:2022,pan:2022,ziyin:2022,simon:2023,cui:2024,raventos:2024,poole:2016,tamai:2023}. For example, during training, AlphaZero~\cite{silver:2018} underwent periods of rapid knowledge acquisition in which increasingly sophisticated chess openings were favored by the engine~\cite{mcgrath:2022}. Large language models (LLMs) have been observed to make sudden improvements in their inductive abilities during training which is related to the formation of special circuitry (so-called \textit{induction heads})~\cite{olsson:2022}. Similar abrupt improvements in specific capabilities, often referred to as breakthroughs, have been observed for a variety of different models and tasks~\cite{ganguli:2022,austin:2021,brown:2020,hendrycks:2020,radford:2021,srivastava:2022,simon:2023,wei:2022,rae:2021,pan:2022,caballero:2022}. Moreover, phenomena such as double descent~\cite{belkin:2019,nakkiran:2021} or grokking~\cite{power:2022,liu:2022,liu2:2022,levi:2023,nanda:2023,rubin:2023} are also reminiscent of phase transitions in physics.

The detection of phase transitions\footnote{In the following, we adopt a more general definition of a phase transition as a sudden shift in the qualitative behavior of a system as a function of a control parameter~\cite{saitta:2011,pan:2022,chen:2023}.} in deep learning systems may improve our understanding and eventually enable better model training. For example, an in-depth analysis of the grokking transition~\cite{power:2022,thilak:2022} led to a way for accelerating generalization~\cite{liu2:2022}. Moreover, it has been shown that models are highly sensitive to perturbations, such as data corruptions, at critical points~\cite{achille:2017,chen:2023}. Being able to predict the behavior of models is also crucial for ensuring safe model deployment~\cite{ganguli:2022} as well as for projecting the performance of future model versions and optimally allocating resources for their training~\cite{kaplan:2020}. 

The characterization of phase transitions in physics is difficult because the state of the systems to be studied typically lives in a very high-dimensional space and is probabilistic in nature, meaning that for given values of the tuning parameters we can find the system in various states. Physicists solve this problem by finding a suitable set of a few low-dimensional quantities, called order parameters~\cite{sethna:2021}, which capture the essence of each phase of the system. For example, even though water is a highly complex system, we can detect the liquid-gas transition by looking at the density which shows a sudden jump at the boiling point and, in this case, serves as an order parameter. However, finding such a suitable set of order parameters is ``considered an art''~\cite{sethna:2021}, as it requires a great deal of human intuition as well as prior understanding. 

Faced with the task of characterizing phase transitions in learning systems based on large NNs, similar issues are encountered. NNs contain an enormous amount of trainable parameters and their state space, as characterized by their neural activations, is huge. This problem is exacerbated in generative models such as LLMs where also the output space is large, i.e., the high dimensionality cannot be foregone by treating the inside of the NN as a black box and focusing solely on its output characteristics. Understanding LLMs from first principles has been notoriously hard~\cite{alishahi:2019}. Theories capturing their microscopic and macroscopic behavior, for instance based on \textit{mechanistic interpretability}~\cite{olah:2022,olsson:2022,wang:2022,nanda:2023,zhong:2023} or \textit{neural scaling laws}~\cite{hestness:2017,rosenfeld:2019,kaplan:2020,henighan:2020,gordon:2021,zhai:2022,hoffmann:2022,caballero:2022}, are still nascent. In particular, the definition of appropriate low-dimensional quantities that facilitate the detection of transitions has been done manually, for example through the extraction of appropriate circuitry~\cite{rauker:2023,conmy:2023,zhong:2023}. Due to this human-in-the-loop, transitions can be easily missed~\cite{zhong:2023} or spuriously induced~\cite{schaeffer:2023}.
 
In physics, these problems have been tackled using statistical methods for the detection of phase transitions from data, which requires minimal prior system knowledge and human input~\cite{wang:2016,carrasquilla:2017,van:2017,wetzel1:2017,wetzel2:2017,zhang:2017,broecker:2017,chng:2017,hu:2017,venderley:2018,carleo:2019,schaefer:2019,rem:2019,greplova:2020,carrasquilla:2020,kottman:2020,arnold:2021,kaming:2021,dawid:2022,arnold:2022,arnold_GEN:2023,arnold_FAST:2023,arnold_FI:2023}. Inspired by this body of work, we here adapt such an approach for the automated detection of phase transitions in LLMs. The method is based on measuring changes in the distribution of the text output of LLMs via generic statistical distances belonging to the family of $f$-divergences, making it {\bf a versatile all-purpose tool for objectively and automatically mapping out phase diagrams of generative models}. Such an approach has the potential to characterize unexplored phase transitions and potentially discover new phases of behaviors. This is crucial in light of the rapid development of LLMs~\cite{achiam:2023,anthropic:2023,team:2023} and their emergent capabilities~\cite{ganguli:2022,austin:2021,brown:2020,hendrycks:2020,radford:2021,srivastava:2022,michaud:2023,olsson:2022,arora:2023,brown:2020,wei:2022,rae:2021,pan:2022}.

As a demonstration, we characterize transitions occurring as a function of three different control parameters in Pythia~\cite{biderman:2023}, Mistral (7B)~\cite{jiang:2023}, and Llama3 (8B)~\cite{llama3modelcard} language models: an integer occurring in the input prompt, the temperature hyperparameter for text generation, and the model's training epoch. 

\paragraph{Specific Findings.} We find that 
\begin{itemize}
\item the instruction-tuned Llama and Mistral models seem to have the capability to order integers whereas all considered base models do not.
\item changes in integer tokenization can be visible in the text output as sharp transitions.
\item three distinct phases of behavior as a function of an LLM's temperature can be mapped out: a deterministic ``frozen'' phase near zero temperature, an intermediate ``coherent'' phase, and a ``disordered'' phase at high temperatures.
\item an LLM's ``heat capacity'' with respect to the temperature can be negative, i.e., the LLM's mean energy can decrease as its temperature is increased.
\item rapid changes in the distribution of weights during training can coincide with transitions in the text output that are present across many prompts.
\item different prompts result in different transition times during training, suggesting that distinct type of behavior can be learned rapidly at distinct times in training.
\end{itemize}

\section{Methodology}\label{sec:methodology}

\subsection{Quantifying Dissimilarity between Distributions} \label{sec:theory}

In this work, we view phase transitions as rapid changes in the probability distribution $P(\cdot|T)$ governing the state of the system $\bm{x} \sim P(\cdot|T)$ as the control parameter $T$ is varied.\footnote{This definition encompasses phase transitions in physics, i.e., abrupt changes in the distribution governing large-scale systems of interacting constituents.} 
That is, values of the parameter at which the distribution changes strongly are considered critical points where phase transitions occur. While it is possible to generalize our approach to distributions conditioned on multiple control parameters (see~\cite{arnold_GEN:2023,arnold_FI:2023}), for simplicity we consider the one-dimensional scenario in the following. 

We quantify the rate of change using $f$-divergences~\cite{liese:2006}, as they have particularly nice properties, such as satisfying the data processing inequality. Given a convex function $f: \mathbb{R}_{\geq 0} \rightarrow \mathbb{R}$ with $f(1)=0$, the corresponding $f$-divergence is a statistical distance defined as
\begin{equation}\label{eq:f_divergence}
    D_{f}[p,q] = \sum_{\bm{x}} q(\bm{x}) f\left(\frac{p(\bm{x})}{q(\bm{x})}\right) \geq 0.
\end{equation}

Prominent examples of $f$-divergences include the Kullback-Leibler (KL) divergence, the Jensen-Shannon (JS) divergence, which corresponds to a symmetrized and smoothened version of the KL divergence, as well as the total variation (TV) distance. Ideally, we would also like the statistical distance we choose to be symmetric $D[p,q] = D[q,p]$. This condition is only satisfied by the TV distance and the JS divergence among the examples above. 

Hence, in this work, we will focus on the TV distance 
\begin{equation}\label{eq:TV_distance}
    D_{\rm TV}[p,q] = \frac{1}{2} \sum_{\bm{x}} |p(\bm{x}) - q(\bm{x})|
\end{equation}
corresponding to $f(x) = \frac{1}{2}|1-x|$, as well as the JS divergence
\begin{equation}\label{eq:JSD}
    D_{\rm JS}[p,q] = \frac{1}{2}D_{\rm KL}\left[p,\frac{p+q}{2}\right] + \frac{1}{2}D_{\rm KL}\left[q,\frac{p+q}{2}\right]
\end{equation}
corresponding to $f(x) = \frac{1}{2}\left[x\log(\frac{2x}{1+x}) +  \log(\frac{2}{1+x})\right]$, where $D_{\rm KL}$ is the KL divergence.

The TV distance and the JS divergence have also had tremendous success in detecting phase transitions in physical systems without prior system knowledge under the name of ``learning-by-confusion''~\cite{van:2017,liu:2018,beach:2018,suchsland:2018,lee:2019,ni:2019,ni2:2019,guo:2020,kharkov:2020,bohrdt:2021,corte:2021,arnold:2022,richter:2022,gavreev:2022,zvyagintseva:2022,zhang:2022,schlomer:2023,guo:2023,arnold_GEN:2023,arnold_FI:2023,arnold_FAST:2023,cohen:2024}.\footnote{Note that both the TV distance and the JS divergence form lower bounds to the KL divergence and other $f$-divergence, such as the $\chi^2$ divergence: $D_{\rm JS}[p,q] \leq D_{\rm TV}[p,q] \leq \sqrt{D_{\rm KL}[p,q]} \leq \sqrt{D_{\chi^2}[p,q]}$~\cite{flammia:2023}. In this sense, detecting a large dissimilarity in terms of the TV distance or the JS divergence also signals a large dissimilarity in other measures.}

\subsection{Detecting Phase Transitions}

Having defined appropriate notions of distance between probability distributions, we now describe their use to detect phase transitions: Consider a sampled set $\mathcal{T}$ of control parameter values $T$, forming a uniform one-dimensional grid. For each $T^{*}$ lying halfway in between grid points, we assess whether it is a critical point by computing a dissimilarity score
$D(T^{*}) = D\left[ P_{\mathrm{left}}(\cdot |T^{*}) , P_{\mathrm{right}}(\cdot |T^{*}) \right]$
between the distributions underlying the segments  $\sigma_{\mathrm{left}}(T^{*})$ to the left and $\sigma_{\mathrm{right}}(T^{*})$ to the right of $T^*$. Denoting the cardinality as $|\cdot|$, we can write these probabilities as $P_{i}(\cdot |T^{*}) = \frac{1}{|\sigma_{i}(T^{*})|} \sum_{ T \in \sigma_i(T^{*}) } P(\cdot | T)$ for $i\in \{ \mathrm{left},\mathrm{right}\}$. Critical points where phase transitions occur can then be identified as local maxima in $D$.

For the sake of simplicity, we proceed with segments of equal length for the rest of this article, and define the length $L=|\sigma_{\rm left}| = |\sigma_{\rm right}|$ as the number of parameter values $T$ to the left or right of $T^{*}$ that characterize the segment. We are free to adjust it according to the problem, as $L$ sets a natural length scale on which changes in the distributions are assessed. Examples will be discussed in Sec.~\ref{sec:results}. In particular for $L=1$ and neighboring parameter points separated by $\delta T$, $D_{f}(T^{*}) = \tfrac12 {f''(1)}\mathcal{F}(T^{*})\delta T^2 + \mathcal{O}(\delta T^3)$ where $\mathcal{F}$ is the Fisher information~\cite{amari:2010}. That is, local changes in a distribution as measured by any $f$-divergence reduce to the Fisher information in the limit $\delta T \to 0$. 
Having the Fisher information as a limiting case is a desirable property: It is a well-known, generic statistical measure for quantifying how sensitive probability distributions are to changes in their parameters and its behavior is well-understood when used to detect phase transitions in physical systems~\cite{you:2007,gu:2010,prokopenko:2011,arnold_FI:2023}.

\subsection{
Application to Language Models and Numerical Implementation}\label{sec:implementation}

In the case of language models, $\bm{x}$ is the sampled text and $T$ is any variable that influences the sampling probability. Because of the autoregressive structure of language models, we can efficiently sample text $\bm{x}$ for a given prompt and evaluate its probability $P(\bm{x}|T)$. Thus, we can obtain an unbiased estimate $\hat{D}(T^{*})$ of $D(T^{*})$ by replacing expected values with sample means where samples correspond to text generated with language models conditioned on different parameter settings $T$, see Appendix~\ref{app:implementation_details} for details on implementation.

For numerical stability and efficient sampling, we express our dissimilarity measures as parameterized by a function $g$ acting on the probability $P(\sigma_{i}|\bm{x}) = \frac {P_{i}(\bm{x})} { P_{\mathrm{left}}(\bm x) +  P_{\mathrm{right}}(\bm x)}$ for $\bm{x}$ to stem from segment $\sigma_{i}$. Specifically, we consider
\begin{equation}
\label{eq:indicator}
   D_g= \frac{1}{2L} \sum_{i \in \{{\rm left}, {\rm right}\}} \sum_{T \in \sigma_{i}} \mathbb{E}_{\bm{x} \sim P(\cdot | T)} \biggl[ g \left[P(\sigma_{i} | \bm{x})\right] \biggr].
\end{equation}
These $g$-dissimilarities and the $f$-divergences [Eq.~\eqref{eq:f_divergence}] defined above correspond to each other in the following sense: any $g$-dissimilarity $D_{g}$ can be rewritten in the form of an $f$-divergence $D_{f}[P_{\rm left}, P_{\rm right}]$ with
\begin{equation}
    f(x) = \frac{x}{2} \cdot g \left(\frac{x}{1+x}\right) + \frac{1}{2}  \cdot g \left(\frac{1}{1+x} \right),
\end{equation}
see Appendix~\ref{sec:app_g_diss} for the derivation and further discussion. In particular, for the choice $g(x) = \log(x) + \log(2)$,  $D_{g}$ corresponds to the JS divergence [Eq.~\eqref{eq:JSD}]. For $g(x) = 1-2 \min \{x, 1-x \}$, $D_{g}$ corresponds to the TV distance [Eq.~\eqref{eq:TV_distance}]. 

A natural choice for $g$ is any linear function in $x$. In particular, setting $g(x) = 2x-1$ results in a dissimilarity measure that quantitifies the ability of an optimal classifier to tell whether a sample $\bm{x}$ has been drawn in the left or right sector. This measure is 0 if the two distributions are completely indistinguishable and 1 if the two distributions are perfectly distinguishable. Moreover, $g(x) = 2x-1$ has the property of being bounded between 1 and -1, where the edge values are attained for the certain predictions 0 and 1, and the value 0 corresponds to uncertain predictions at 0.5. This results in a low variance and favorable convergence properties for $\hat{D}_{g(x)=2x-1}$, which we will refer to as \textit{linear dissimilarity} in what follows. This quantity is a valid $f$-divergence and reduces to the Fisher information in lowest non-vanishing order\footnote{In fact, any $g$-dissimilarity with $g(1/2)=0$ and a twice-differentiable $g$-function can be shown to be proportional to the Fisher information in lowest order.}, as shown in Appendix~\ref{sec:app_g_diss}.

\subsection{Utilized Large Language Models}\label{sec:utilized_LLMs}

In this work, we study transitions in models of the Pythia,  Mistral, and Llama family. 

\paragraph{Pythia} Pythia~\cite{biderman:2023} is suite of 16 LLMs released in 2023 that were trained on public data in the same reproducible manner ranging from 70 million (M) to 12 billion (B) parameters in size. We consider every second model, i.e. the models with 70M, 410M, 1.4B, and 6.9B parameters.

\paragraph{Mistral} From the Mistral family, we consider the base model Mistral-7B-v0.1 with 7.3B parameters and the corresponding fine-tuned Mistral-7B-Instruct model~\cite{jiang:2023} released in 2023.

\paragraph{Llama}
Llama 3~\cite{llama3modelcard} from Meta AI was released in 2024. We consider both the Llama-3 8B parameter base model and NVIDIA's chat-tuned Llama3-ChatQA-1.5-8B~\cite{liu:2024}. For the chat model we use accordingly formatted inputs.

\section{Results}\label{sec:results}
In the following, we will explore all three fundamental ways in which a parameter $T$ may influence the output distribution of a language model: $(i)$ As a variable within the input prompt, we scan through integers injected to the prompt in Sec.~\ref{sec:prompt_scan}. $(ii)$ As a hyperparameter controlling how a trained language model is applied, we vary the temperature in Sec.~\ref{sec:temperature_scan}. $(iii)$ As a training hyperparameter of the language model, we vary the number of training epochs in Sec.~\ref{sec:epoch_scan}.

\subsection{Transitions as a Function of a Variable in the Prompt}
\label{sec:prompt_scan}

\begin{figure}[htb]
	\centering
  	\includegraphics[width=0.49\linewidth]{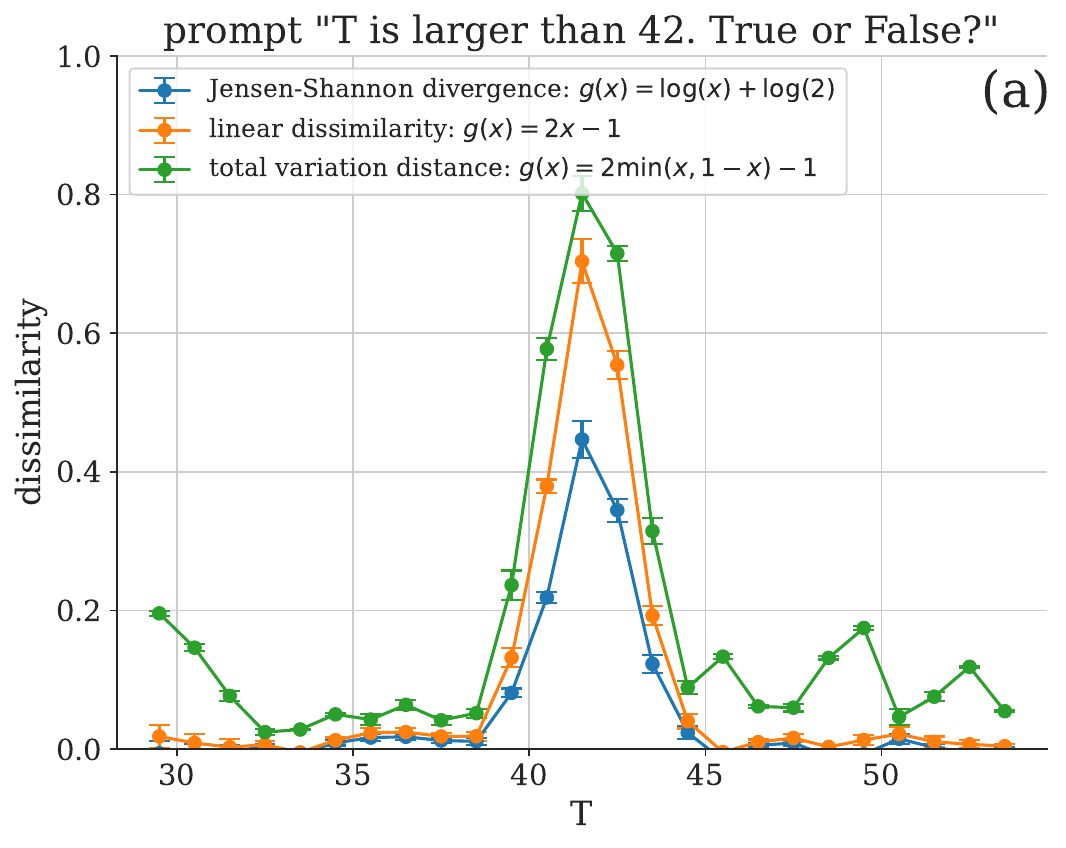}		
     	\includegraphics[width=0.49\linewidth]{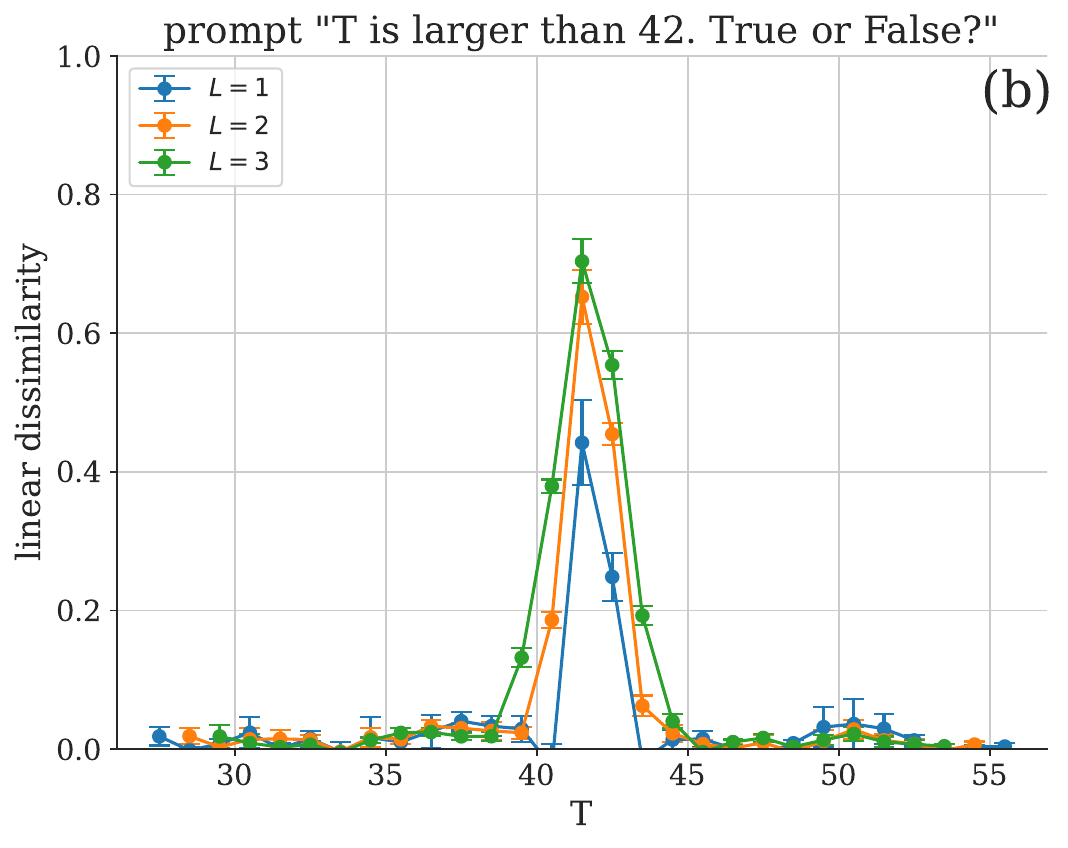}

		\caption{Mistral model applied to the integer ordering prompt. (a) Different $g$-dissimilarities with $L=3$. (b) Linear dissimilarity for different $L$-values. [Number of text outputs generated per parameter value $T$: 10280. Number of generated output tokens: 10. Error bars indicate standard error of the mean over 4 batches, each with batch size 2056.]}
		\label{fig:prompt_token_mistral}
\end{figure}

As an introduction, we start with the simplest case: The parameter $T$ to be varied is a particular part of the prompt, and all parameters of the language model itself are fixed. As a first such prompt, consider \textit{``$T$ is larger than 42. True or False?''} with an integer $T$ as the control parameter. An LLM that understands the order of integers should output very different answers for $T<42$ versus $T>42$, i.e., its distribution over outputs should change drastically around $T=42$. Thus, in such a case we expect the dissimilarities to show a clear peak around $T=42$.

Figure~\ref{fig:prompt_token_mistral}(a) shows dissimilarities based on various $g$-functions for the Mistral-7B-Instruct model~\cite{jiang:2023}. All dissimilarities show a clear peak around $T=42$, whereas they are relatively flat otherwise. This is a clear example of an abrupt transition between two distinct phases of behaviors of an LLM as a function of a tunable parameter. As compared to the linear dissimilarity, the logarithm-based JS divergence is arguably a bit sharper in that it decays more rapidly to baseline 0. The TV distance's peak is the broadest due to the $\min$ function appearing in its $g$-function. In the following, we will focus on the linear dissimilarity as a compromise between sensitivity and numerical stability.

The transition is also clearly visible using different $L$ settings, see Fig.~\ref{fig:prompt_token_mistral}(b). Smaller $L$ values are closer to the Fisher information limit, while larger values generally lead to higher distinguishability of distributions and therefore larger peaks at transition points. As we will see in more detail in Sec.~\ref{sec:epoch_scan}, they can also be less susceptible to outliers due to the averaging over several parameter points.

\begin{figure}[htb]
	\centering

    \includegraphics[width=0.49\linewidth]{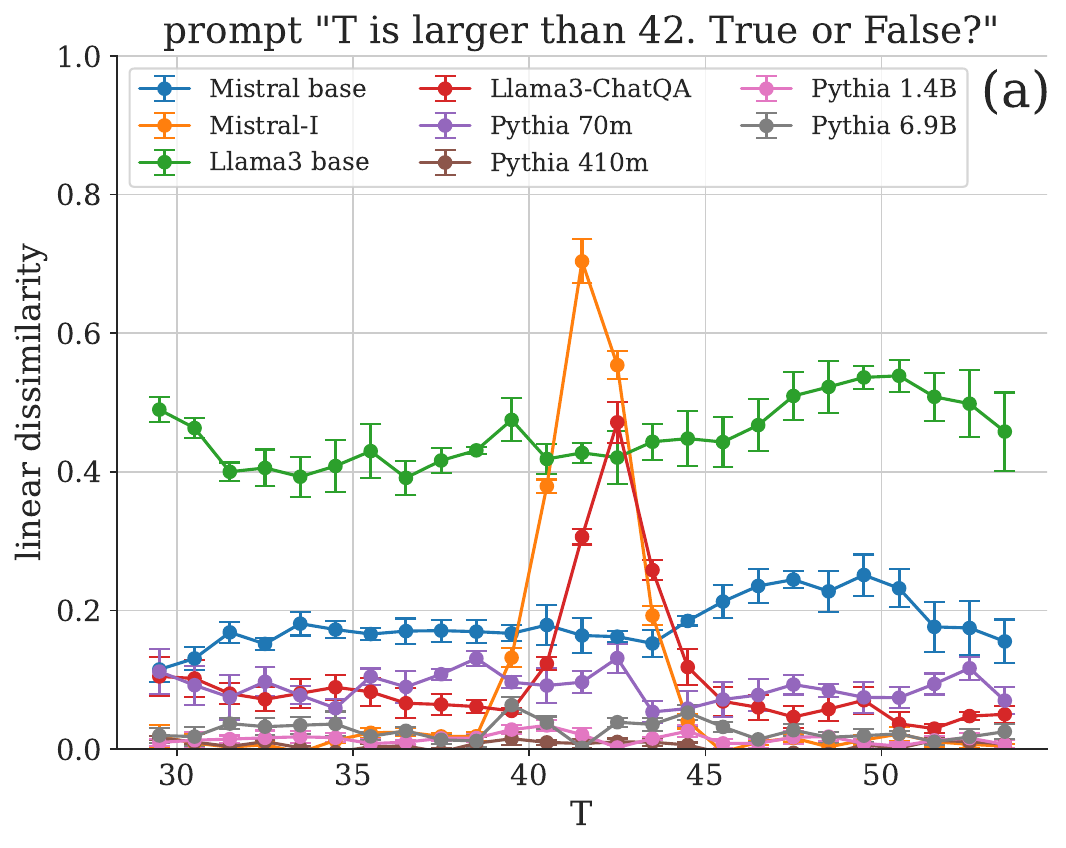}
    \includegraphics[width=0.49\linewidth]{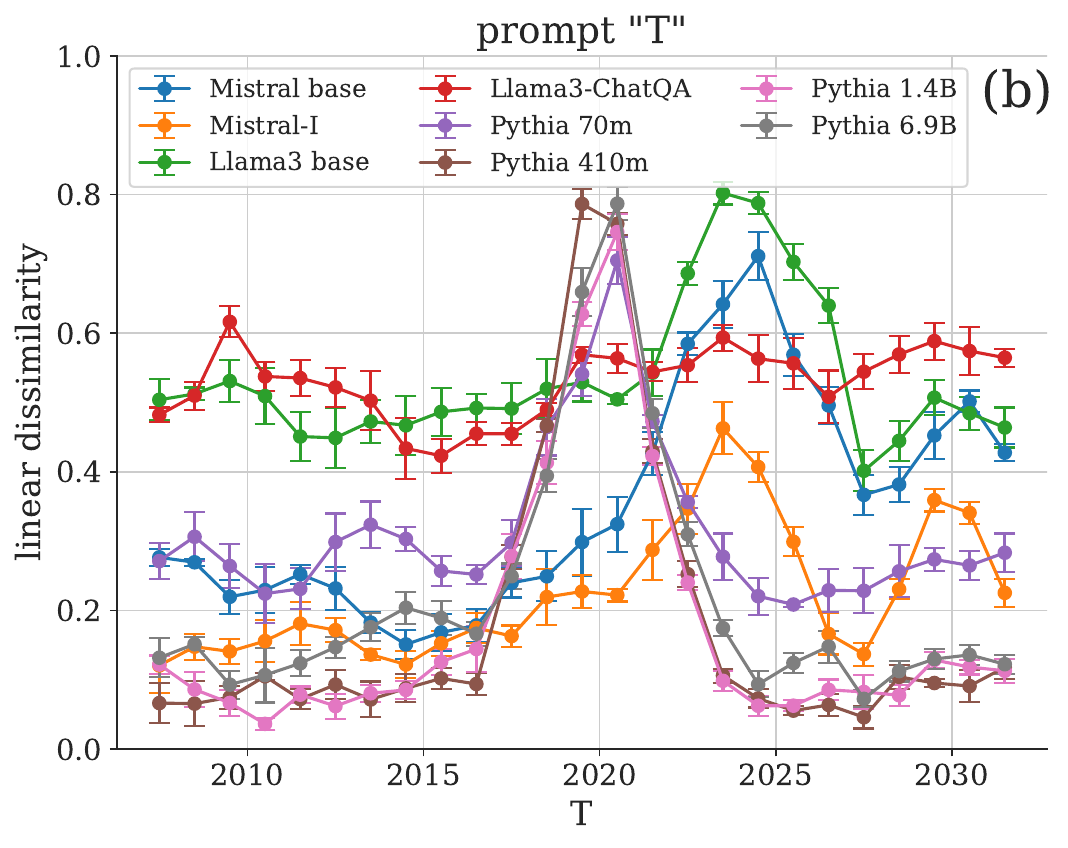}
    
		\caption{Benchmarking of various models using the linear dissimilarity with $L=3$.  (a) Test of ability to compare integers in value. (b) Bare integers as prompt reveals transition in tokenizer encoding. [Same numerical settings as in Fig~\ref{fig:prompt_token_mistral}.]
  }
		\label{fig:prompt_token_all}
\end{figure}

Interestingly, when performing the same analysis on base models such as the Llama3-8B and Mistral base models, as well as Pythia models~\cite{biderman:2023} of various sizes, the resulting linear dissimilarity is flat, signaling the absence of any transition [see Fig.~\ref{fig:prompt_token_all}(a)]. In contrast to Mistral-7B-Instruct and NVIDIA's chat-tuned Llama3-8B, these models do not show a clear peak around $T=42$.

A transition of a different origin can be observed in Fig.~\ref{fig:prompt_token_all}(b), where the LLMs are probed using the prompt \textit{``$T$''} with $T$ again being an integer.
Interestingly, all Pythia models show a peak between $T=2020$ and $T=2021$. This behavioral transition may originate from a transition in the tokenizers of these models, which encode numbers in a range below $T=2021$ with a single token and numbers in a range at and above $T=2021$ with two. This explanation is corroborated by the absence of the transition around $T=2021$ for the Llama and Mistral models, whose tokenizers translate a number into tokens following rules that are independent of the number's frequency.

The Mistral models and the base Llama3-8B model show a smaller peak around $T=2023/2024$. Both models have only encountered training data from before and around that time given their release date in 2023/2024, which may explain the peak. This transition is absent in the Pythia models.

\subsection{Transitions as a Function of the Model's Temperature}
\label{sec:temperature_scan}

Next, we consider transitions as a function of the temperature hyperparameter $T$ controlling how the logits $\bm{z}$ are converted to probabilities 
\begin{equation}
    p_{i} = \frac{e^{z_{i}/T} }{\sum_{j} e^{z_{j}/T}} 
    \label{eq:p_from_logits}
\end{equation}
for next-token prediction where the sum runs over all possible tokens. Per construction, at $T=1$ language models predict probabilities $p_i$ to approximate the distribution to be learned. In the limit $T \rightarrow 0$, the model deterministically picks the most likely next token in each step. For $T \rightarrow \infty$ the model samples the next token uniformly.

This scenario somewhat resembles a system of a one-dimensional lattice of spins that are coupled via long-range interactions, i.e., the one-dimensional Ising model~\cite{dyson:1968,martinez:2022}, which has an order-disorder phase transition. In our case, the tokens take the role of the spins, and the coupling is mediated via the transformer's attention mechanism.

\begin{figure}[htb!]
	\centering
     \includegraphics[width=0.49\linewidth]{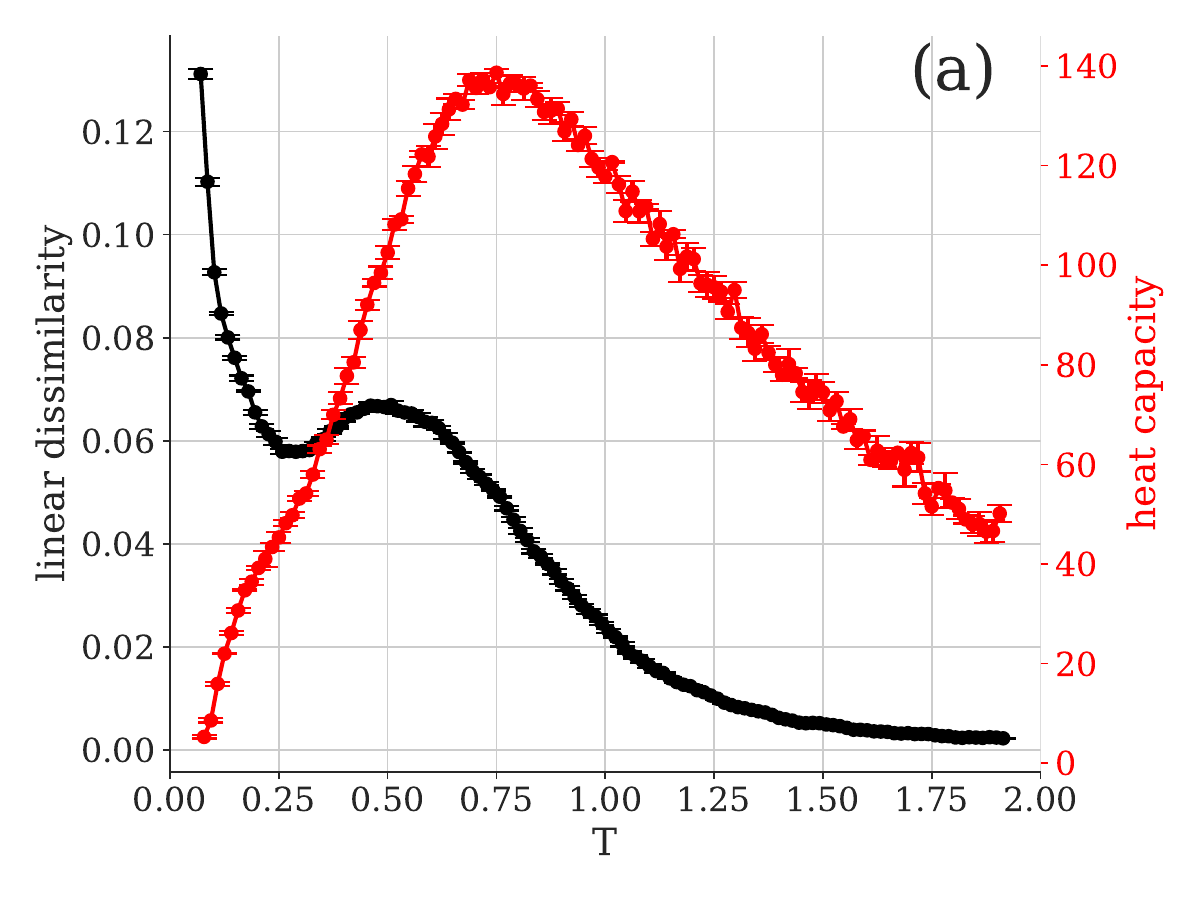}
    \includegraphics[width=0.49\linewidth]{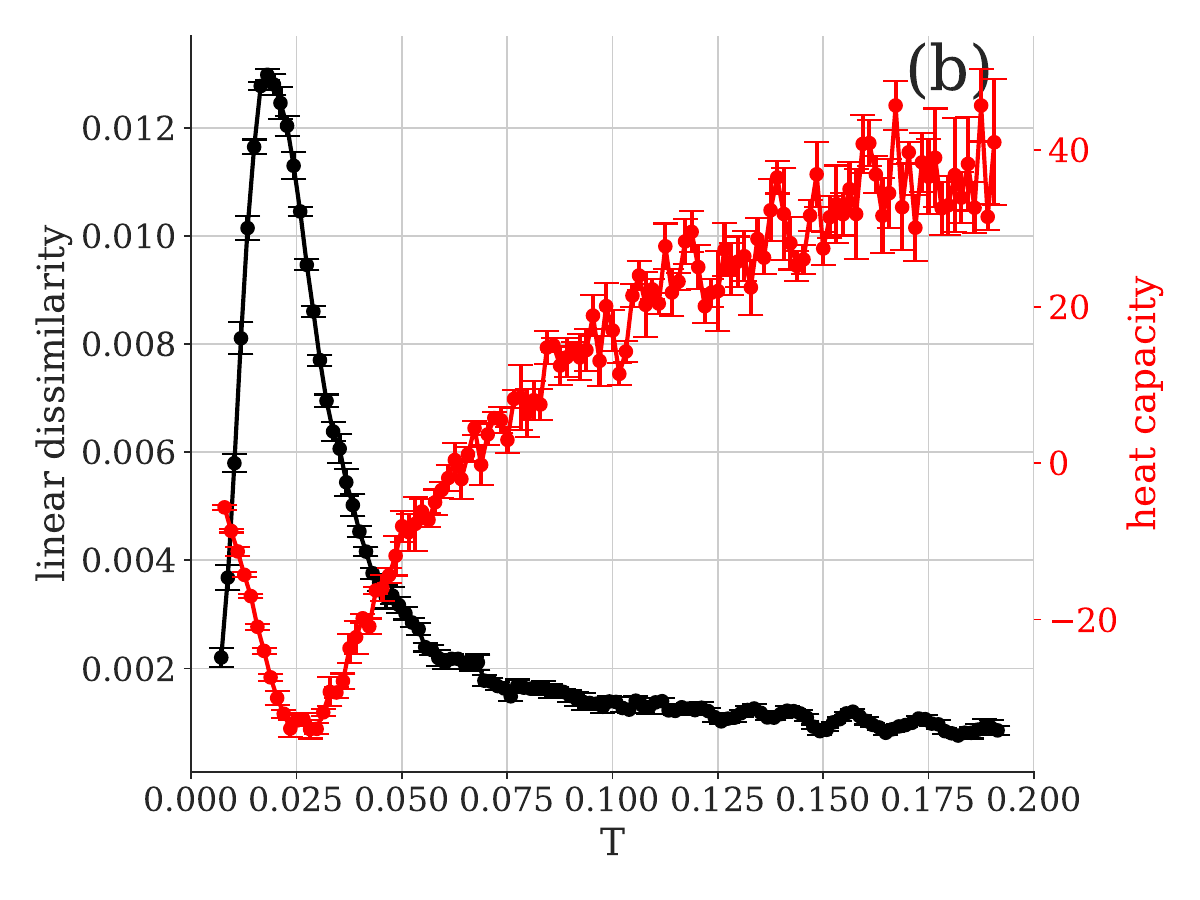}
  \caption{Temperature transitions of Pythia 70M model in response to the prompt ``There’s measuring the drapes, and then there’s measuring the drapes on a house you haven’t bought, a'' -- an excerpt from OpenWebText~\cite{gokaslan:2019}. Linear dissimilarity measure ($L=5$) is shown in black. Heat capacity is shown in red. Dashed lines indicate local maxima, i.e., predicted critical points. Shaded regions indicate the error bands. (a) Temperature range $[10^{-4}, 2]$. (b) Zoomed-in range $[10^{-4}, 0.2]$ near $T=0$. [Number of text outputs generated per parameter value $T$: 20480. Number of generated output tokens: 10. Error bars indicate standard error of the mean over 4 batches, each with batch size 5120.]}
		\label{fig:temperature}

\end{figure}

In Fig.~\ref{fig:temperature}, the dissimilarity shows two distinct peaks corresponding to two transition points: one at a very low temperature $T_{1}^{*} \approx 0.02$ and one at an intermediate temperature $T_{2}^{*} \approx 0.5$. Intuitively, these two points mark transitions between three distinct phases of behavior: ``frozen'' at low temperatures $T < T_{1}^{*}$, ``unfrozen and sensible'' at intermediate temperatures $T_{1}^{*} < T < T_{2}^{*}$, and ``random'' at high temperatures $T_{2}^{*} < T$. The transition at low temperatures has recently been investigated in Ref.~\cite{bahamondes:2023} for GPT-2 using physics-inspired quantities. Moreover, they speculated on the existence of a phase transition at higher temperatures. 

We perform an analysis independent of the dissimilarity-based indicators by taking inspiration from statistical mechanics, where the state of thermal systems is governed by the Boltzmann distribution (see Appendix~\ref{app:energy_based} for details). We view the LLM as such a thermal system at varying temperature where the negative logarithmic probability at $T=1$, $- \log P(\bm{x}|T=1)$, takes on the role of the energy $E$ of a given text output $\bm{x}$. In physical systems governed by Boltzmann distributions, thermal phase transitions can be detected as peaks in the heat capacity $C(T) = \partial \mathbb{E}_{\bm{x} \sim P(\cdot|T)}\left[E(\bm{x})\right]/\partial T$, i.e., by looking at the temperature derivative of the mean total energy~\cite{blundell:2009}. 

Figure~\ref{fig:temperature} shows that the locations of peaks (i.e., dips) in these quantities are close to the critical points highlighted by our method. Note that in the LLM case, the text outputs are not truly sampled from a Boltzmann distribution governed by the total energy. Instead, each individual token is drawn from a Boltzmann distribution for its individual energy conditioned on the previous tokens only. This procedure corresponds to a greedy sampling strategy. The resulting sampling mismatch can lead to the counterintuitive phenomenon of the mean energy of the system increasing with decreasing temperature corresponding to a negative ``heat capacity'', cf. Fig.~\ref{fig:temperature}(b).

The intermediate temperature transition at $T_{2}^{*}$ may be reminiscent of the Schottky anomaly~\cite{blundell:2009} occurring in systems with a finite number of energy levels. As such, this phenomenon is perhaps better described as a crossover rather than a phase transition in the Ehrenfest sense. In particular, we also observed such a transition for a basic language model that samples words according to their overall frequency without taking into account any word-to-word interaction. 

In Fig.~\ref{fig:temperature} we have investigated the output distributions corresponding to a specific prompt. While we find that the temperature behavior is strongly dependent on the prompt, there seems to be a trend: many distinct prompts lead to a transition at $T \approx 1$ (i.e., on the order of the natural temperature scale), at $T \ll 1$, or both.

\subsection{Transitions as a Function of the Training Epoch}
\label{sec:epoch_scan}

Finally, we search for transitions as a function of the training epoch, i.e., we compare the output distributions of models at different stages during training and see whether there are certain epochs at which these statistics change drastically. Such temporal analyses are rare given that they require access to models at checkpoints during training~\cite{liu:2021,gurnee:2023,chen:2023}. Here, we analyze the Pythia suite of models for which such checkpoints are publicly available. 

Ref.~\cite{millidge:2023} analyzed the weight distribution of the Pythia models, and similar weight-based analyses of other NNs during training have also been performed in previous works~\cite{shwartz:2017,achille:2017,chen:2023}. In order to study the previously observed transitions~\cite{millidge:2023}, 
we analyze changes in the weight distributions in the same manner as for the output distributions (see Sec.~\ref{sec:methodology}), i.e., to characterize phase transitions using dissimilarities. The lists of model weights are converted to distributions via histogram binning (10000 bins for the range -3 to 3).

\begin{figure}[htb!]
  		\includegraphics[height=0.31\linewidth]{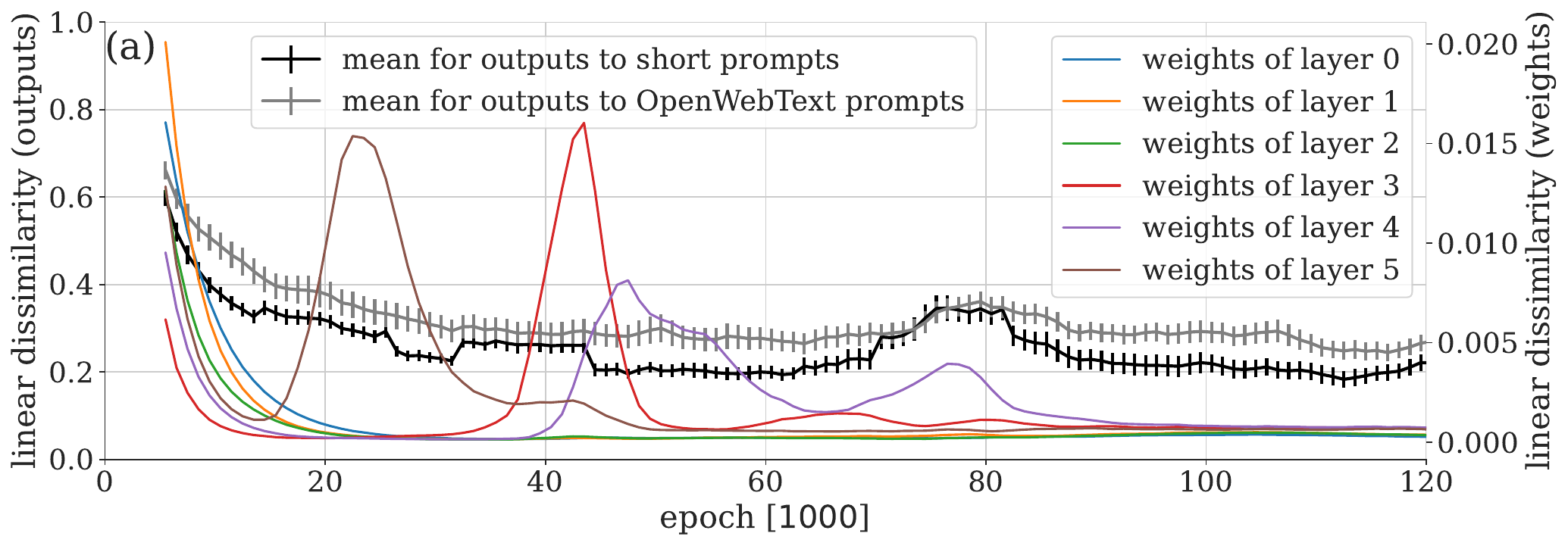}	
  		\includegraphics[height=0.31\linewidth]{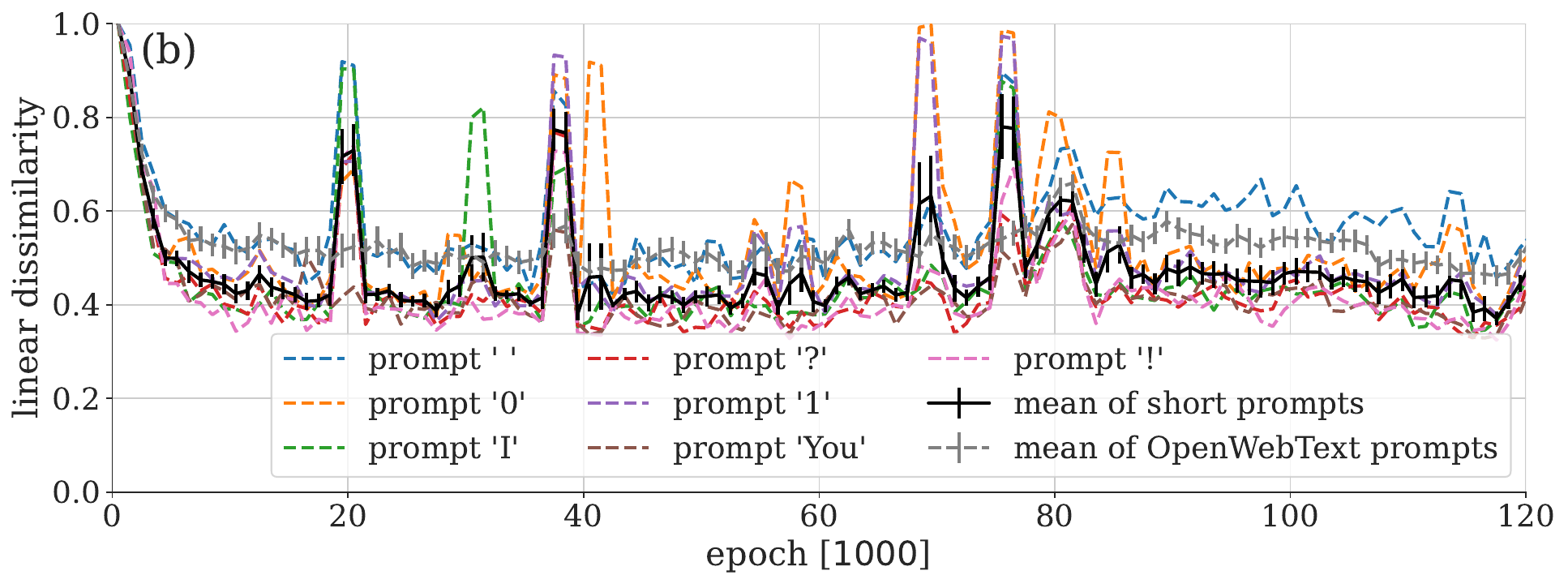}	
    
		\caption{Linear dissimilarity by epoch, with checkpoints taken every 1000 epochs. (a) Computed at $L=6$ for both weights and responses to 20 random prompts from OpenWebText (gray) and 7 short prompts (black) shown in panel (b). (b) Computed at $L=1$ for several prompts. For reference, the mean linear dissimilarity over short prompts and OpenWebText prompts with $L=1$ is also shown. [Number of text outputs generated per parameter value $T$ and prompt: 1536. Number of generated output tokens: 10. Error bars indicate the standard error of the mean over all corresponding prompts. Error bars for the individual prompts in panel (b) are almost negligible and thus omitted to avoid visual clutter.]
  }
		\label{fig:epochs}
\end{figure}

The results for $L=6$ are shown in Fig.~\ref{fig:epochs}(a) as colored lines, each corresponding to the distribution of the weights of a particular QKV layer. Different layers show transitions at roughly 20K (layer 5), 40K (layers 3), 50K (layer 4), and 80K (layer 4) epochs. We also observe a large peak around epoch 0, i.e., at the start of the training, highlighting that the LLM learns most rapidly at the beginning stages. In the long run, the dissimilarity curves approach 0, signaling that overall the weight distributions become less and less distinguishable.

Complementarily, in the same plot, we show dissimilarities derived from the LLM output distributions. The grey line corresponds to an average of the dissimilarities obtained by using entries from OpenWebText~\cite{gokaslan:2019} (which serves as a proxy for the Pythia training dataset) as prompts. The black line corresponds to the average of results obtained from a selection of single-token prompts [see also panel (b)]. Both dissimilarity curves show a peak around epoch 0 as well as a peak around 80K epochs that is potentially related to the rapid change of layer 4 around the same time.

Figure~\ref{fig:epochs}(b) shows dissimilarities as a function of the training epoch for models queried with short, generic prompts (`` '', ``0'', ``I'', ``?'', ``1'', ``You'', and ``!'') at $L=1$. These short prompts were selected to be as general as possible and the associated output distributions seem more sensitive as compared to the long examples from OpenWebText: their mean dissimilarity shows clear peaks near epochs 20K, 40K, and 80K. These correspond to outliers where the output distribution changes severely only at a single point and returns back (close) to its original behavior immediately after. As such, these peaks do not mark transitions between two macroscopic phases of behavior. We further verified this outlier behavior by inspecting the dissimilarity between the points directly to the left and right of the potential outlier. It remains an open question if these outliers are linked to the transitions observed in the layer weights shown in panel (a). 

The larger $L$ value used in panel (a) averages out the signal stemming from these outliers. Such a reduced susceptibility to outliers can be an advantage of using $L \gg 1$ when searching for macroscopic transitions in particular. 

Some peak locations in the dissimilarity curves are prompt-dependent, indicating that learning progresses differently for different types of behavior. 
Here we have used rather generic prompts, resulting in an analysis of the LLM's general behavior during training. However, in principle, conditioning on the prompt allows one to analyze whether and when specific knowledge emerges~\cite{liu:2021,gurnee:2023}. As an outlook, one can imagine automatically monitoring changes across a multitude of prompts on different topics and testing different abilities at scale, without the need to design individual metrics for each prompt.

\section{Related Works}\label{sec:discussion}
Before concluding, let us discuss how our method relates to other approaches for studying transitions in LLMs.

{\bf Generic performance-based analysis.} Many previous works found transitions in LLM behavior by locating sharp changes in generic performance measures, such as sudden drops in training loss~\cite{olsson:2022,chen:2023}. While this may capture transitions in the overall behavior, such an approach cannot resolve transitions in specific LLM behavior. In particular, it may miss algorithmic transitions where the same performance is reached but by different means~\cite{zhong:2023}.

{\bf Prompt-specific success metrics.} Other works have found transitions by looking at success metrics tailored toward specific prompts~\cite{austin:2021,brown:2020,hendrycks:2020,radford:2021,srivastava:2022,wei:2022,liu:2021}. Recalling the example studied in Sec.~\ref{sec:prompt_scan}, this would correspond to assigning a score of 1 if the LLM provided the correct answer to the question $x < 42$ and 0 otherwise. Similarly, one could compute such a score in a temporal analysis (Sec.~\ref{sec:epoch_scan}) or for detecting transitions as a function of another hyperparameter (Sec.~\ref{sec:temperature_scan}). A downside of this approach is that it is restricted to prompts that allow for a clear score to be assigned. In particular, choosing an appropriate scoring function may require lots of human engineering. Moreover, discontinuous metrics can artificially induce transitions where the underlying behavior varies smoothly~\cite{schaeffer:2023}. Similarly, they may miss transitions where the same performance is reached but by different means~\cite{zhong:2023}.

{\bf Measures based on model internals.} The aforementioned approaches are based on the model output. Many works have also detected transitions based on changes in the internal structure of models, such as its trainable parameters~\cite{millidge:2023,chen:2023} (similar to the weight-based analysis we have performed in Sec.~\ref{sec:epoch_scan}). However, access to model internals may not always be available. Moreover, the design of measures that capture specific transitions in behavior requires lots of human input~\cite{rauker:2023,conmy:2023,zhong:2023}, e.g., using insights from the field of \textit{mechanistic interpretability}.

\section{Conclusion and Outlook}
\label{sec:conclusion}
We have proposed a method for automating the detection of phase transitions in LLMs, and demonstrated that it successfully reveals a variety of transitions. Leveraging access to the LLMs' next-token probability distributions, the proposed dissimilarity measures can efficiently quantify distribution shifts without fine-tuning or adaption to the specific scenario at hand. Because the method is solely based on analyzing a model's output distribution and access to the model weights is not required, it enables \textit{black-box interpretability} studies.

The proposed method is not only applicable to language models, but can be straightforwardly adapted to any generative model with an explicit, tractable density~\cite{goodfellow2:2016,arnold_GEN:2023}. If one can draw samples from the output distribution but does not have explicit access to the underlying probabilities, then the dissimilarity measures can still be approximated using NN-based classifiers~\cite{menon:2016,arnold_FI:2023} tailored toward the particular data type, such as natural language.

\paragraph{Limitations.} Future large-scale investigations are needed to fully understand how the uncovered transitions depend on variables such as the specific prompt, the number of generated output tokens, or the selected model. In particular, due to computational resource constraints, the size of the studied language models has been limited.

\paragraph{Broader Impact.} Our method has the potential to enhance the development of future AI systems due to an improved understanding of their behavior. The dual-use nature of such systems carries inherent risks, which requires one to proceed with caution and implement mechanisms to ensure they are used safely and ethically.

\begin{ack}
We thank Christoph Bruder for stimulating discussions and helpful suggestions on the manuscript. J.A. and N.L. acknowledge financial support from the Swiss National Science Foundation individual grant (grant no. 200020 200481). This material is based upon work supported by the National Science Foundation under grant no. OAC-1835443, grant no. OAC-2103804, and grant no. DMS-2325184. The authors acknowledge the MIT SuperCloud and Lincoln Laboratory Supercomputing Center for providing HPC resources that have contributed to the research results reported within this paper.
\end{ack}

\newpage
\bibliographystyle{unsrt}
\bibliography{refs.bib}

\appendix
\newpage

\setcounter{equation}{0}
\setcounter{figure}{0}
\setcounter{table}{0}
\makeatletter
\renewcommand{\thesection}{A}
\renewcommand{\theequation}{A\arabic{equation}}
\renewcommand{\thefigure}{A\arabic{figure}}
\section{Implementation details}\label{app:implementation_details}

\paragraph{Estimating dissimilarity measures.} Our method for detecting phase transitions is based on estimating $D_{g}$ across the entire parameter range. Starting with a fixed set of points on a uniform grid $\mathcal{T}$, let us denote the set of in-between points at least $L$ points away from the border of the range as $\bar{\mathcal{T}}$ (note that $|\bar{\mathcal{T}}| = |\mathcal{T}|-2L$). For each trial point $T^{*}\in \bar{\mathcal{T}}$, we obtain an unbiased estimate $\hat{D}_{g} = (\hat{J}_{\rm left} + \hat{J}_{\rm right})/2$, where 
\begin{equation}\label{eq:estimate}
   \hat{J}_{i}= \frac{1}{|\sigma_{i}|} \sum_{T \in \sigma_{i}} \frac{1}{\mathcal{D}(T)}\sum_{\bm{x} \in  \mathcal{D}(T)} g \left[P(\sigma_{i} | \bm{x})\right].
\end{equation}
Recall that $\sigma_{\rm left}$ and $\sigma_{\rm right}$ denote the $L$ closest points to the left or right of the trial point. Here, $\mathcal{D}(T)$ denotes a set of output texts $\bm{x}$ generated via the LLM at point $T \in \mathcal{T}$. In this work, we choose the number of generated text samples to be the same for all $T\in \mathcal{T}$, i.e., $|\mathcal{D}(T)| = |\mathcal{D}|$. 

We perform the computation of $\hat{D}_{g}(T^{*})\;\forall \,T^{*} \in \bar{\mathcal{T}}$ in two stages. In a first stage, we go through each grid point $T \in \mathcal{T}$ and generate text outputs that are $N_{\rm tokens}$ in length via the LLM. The associated computation time scales as $|\mathcal{T}|\cdot N_{\rm tokens} \cdot t_{\rm LLM,eval}(|\mathcal{D}|)$, where $t_{\rm LLM,eval}(|\mathcal{D}|) = \mathcal{O}(|\mathcal{D}|)$ corresponds to the time it takes the LLM to generate $|\mathcal{D}|$ different outputs (single token in length). In a second stage, for a given trial point $T^{*} \in \bar{\mathcal{T}}$, we evaluate the probability of each text output generated in its vicinity $\{ \bm{x} \in \mathcal{D}(T) | T \in \sigma_{\rm left}(T^{*}) \cup \sigma_{\rm right}(T^{*}) \}$ to come from the left or right segment. That is, we compute $P(\bm{\sigma}_{\rm left}|\bm{x})$ and $P(\bm{\sigma}_{\rm right}|\bm{x})$, i.e., a term in the sum of Eq.~\eqref{eq:estimate}. The computation time associated with the second stage scales as $|\bar{\mathcal{T}}| \cdot N_{\rm tokens} \cdot t_{\rm LLM,eval}(|\mathcal{D}|) \cdot 2L$. Note that in practice, one can embarrasingly parallelize over the different grid and trial points. Moreover, one generates and evaluates text outputs batchwise.

\paragraph{Compute Resources.} For our computations, we used an NVIDIA RTX 3090 GPU as well as NVIDIA Tesla V100 (32 GB) and NVIDIA A100 (40GB) GPUs.

\paragraph{Code availability.} A \texttt{Python} implementation of our method is available at \href{https://github.com/llmtransitions/llmtransitions}{github.com/llmtransitions/llmtransitions}.

\paragraph{Utilized assets.} Our code is implemented in \texttt{Python} and internally uses NumPy~\citeSI{harris:2020}, PyTorch~\citeSI{pytorch:2019}, and transformers~\citeSI{wolf:2020}. For the presented examples, we make use of the following additional packages: pandas~\citeSI{mckinney:2010}, SciPy~\citeSI{scipy:2020}, Matplotlib~\citeSI{hunter:2007}, and seaborn~\citeSI{waskom:2021}.

The \href{https://huggingface.co/collections/EleutherAI/pythia-scaling-suite-64fb5dfa8c21ebb3db7ad2e1}{Pythia models}, the Mistral models \href{https://huggingface.co/mistralai/Mistral-7B-Instruct-v0.1}{Mistral-7B-Instruct model} and
\href{https://huggingface.co/mistralai/Mistral-7B-v0.1}{Mistral-7B-Base model}, are available under the Apache-2.0 license on Hugging Face. The \href{https://huggingface.co/meta-llama/Meta-Llama-3-8B}{Meta-Llama-3-8B} model and NVIDIA's  \href{https://huggingface.co/nvidia/Llama3-ChatQA-1.5-8B}{
Llama3-ChatQA-1.5-8B} are available on Hugging Face under the Meta Llama 3 community license.

\setcounter{equation}{0}
\setcounter{figure}{0}
\setcounter{table}{0}
\makeatletter
\renewcommand{\thesection}{B}
\renewcommand{\theequation}{B\arabic{equation}}
\renewcommand{\thefigure}{B\arabic{figure}}
\section{Theoretical background on $g$-dissimilarities}\label{sec:app_g_diss}

\paragraph{Correspondence Between $f$-divergences and $g$-dissimilarities.} Let us establish a correspondence between $f$-divergences [Eq.~\eqref{eq:f_divergence}] and $g$-dissimilarities [Eq.~\eqref{eq:indicator}]. We can write any $g$-dissimilarity as
\begin{align*}
    D_{g} &= \frac{1}{2}\left(\mathbb{E}_{\bm{x} \sim P_{\rm left}} \biggl[ g \left[P(\sigma_{\rm left} | \bm{x})\right] \biggr] + \mathbb{E}_{\bm{x} \sim P_{\rm right}}\biggl[ g \left[P(\sigma_{\rm right} | \bm{x})\right] \biggr] \right)\\
    &= \mathbb{E}_{\bm{x} \sim P_{\rm right}} \biggl[\frac{1}{2}\frac{P_{\rm left}(\bm x)}{P_{\rm right}(\bm x)} g \left[P(\sigma_{\rm left} | \bm{x})\right] \biggr] + \mathbb{E}_{\bm{x} \sim P_{\rm right}}\biggl[ \frac{1}{2}g \left[P(\sigma_{\rm right} | \bm{x})\right] \biggr]\\
&= \mathbb{E}_{\bm{x} \sim P_{\rm right}} \biggl[\frac{1}{2}\frac{P_{\rm left}(\bm x)}{P_{\rm right}(\bm x)}g \left[P(\sigma_{\rm left} | \bm{x})\right] + \frac{1}{2}g \left[P(\sigma_{\rm right} | \bm{x})\right] \biggr] \\
&= \mathbb{E}_{\bm{x} \sim P_{\rm right}} \biggl[\frac{1}{2}\frac{P_{\rm left}(\bm x)}{P_{\rm right}(\bm x)}g \left[ 
    \frac{P_{\rm left}(\bm{x})} { P_{\mathrm{left}}(\bm x) +  P_{\mathrm{right}}(\bm x)} 
    \right] + \frac{1}{2}g \left[
\frac{P_{\rm right}(\bm{x})} { P_{\mathrm{left}}(\bm x) +  P_{\mathrm{right}}(\bm x)}
    \right] \biggr] \\
    &= \mathbb{E}_{\bm{x} \sim P_{\rm right}} \biggl[\frac{1}{2}\frac{P_{\rm left}(\bm x)}{P_{\rm right}(\bm x)}g \left[ 
    \frac{\frac{P_{\rm left}(\bm{x})}{P_{\mathrm{right}}(\bm x)}} { \frac{P_{\mathrm{left}}(\bm x)}{P_{\mathrm{right}}(\bm x)} +  1} 
    \right] + \frac{1}{2}g \left[
\frac{1} { \frac{P_{\mathrm{left}}(\bm x)}{P_{\mathrm{right}}(\bm x)} +  1}
    \right] \biggr]. \\
\end{align*}
Thus, any $g$-dissimilarity $D_{g}$ can be rewritten in the form of an $f$-divergence $D_{f}[P_{\rm left}, P_{\rm right}]$ with
\begin{equation}
    f(x) = \frac{x}{2} \cdot g \left(\frac{x}{1+x}\right) + \frac{1}{2}  \cdot g \left(\frac{1}{1+x} \right).
\end{equation}
Note, however, that not any choice of $g$-function will lead to a proper $f$-divergence in the sense that the resulting $f$-function may not be convex and $f(1)$ may not be zero (recall the definition of an $f$-divergence in Sec.~\ref{sec:theory}).
\paragraph{JS divergence.}
Using the correspondence above, we have that $D_{g(x)= \log(x) + \log(2)}$ is equivalent to an $f$-divergence $D_{f}[P_{\rm left}, P_{\rm right}]$ with
\begin{equation}
    f(x) = \frac{x}{2} \cdot \log \left(\frac{2x}{1+x}\right) + \frac{1}{2}  \cdot \log \left(\frac{2}{1+x} \right) 
\end{equation}
which corresponds to the JS divergence, see Eq.~\eqref{eq:JSD}. 

\paragraph{TV distance.} Let us further prove that the $D_{g(x)= 1 - 2\min \{ x,1-x \} }$ corresponds to the TV distance $D_{\rm TV}[P_{\rm left}, P_{\rm right}]$. We have
\begin{align}\label{eq:TV_proof_1}
    D_{g(x)= 1 - 2\min \{ x,1-x \} } = 1 &- \mathbb{E}_{\bm{x} \sim P_{\rm left}} \biggl[ \min \{ P(\sigma_{\rm left} | \bm{x}), P(\sigma_{\rm right} | \bm{x}) \} \biggr] \nonumber\\
    &- \mathbb{E}_{\bm{x} \sim P_{\rm right}}\biggl[ \min \{ P(\sigma_{\rm left} | \bm{x}), P(\sigma_{\rm right} | \bm{x}) \}  \biggr] 
\end{align}
Using the identity 
\begin{equation*}\label{eq:TV_proof_2}
    \min \{ P(\sigma_{\rm left} | \bm{x}), P(\sigma_{\rm right} | \bm{x}) \} = \frac{1}{2} \left( 1 - \left|P(\sigma_{\rm left} | \bm{x}) -  P(\sigma_{\rm right} | \bm{x})\right|\right),
\end{equation*}
Eq.~\eqref{eq:TV_proof_1} can be rewritten as 
\begin{align*}\label{eq:TV_proof_3}
    D_{g(x)= 1 - 2\min \{ x,1-x \} } &= \frac{1}{2} \mathbb{E}_{\bm{x} \sim P_{\rm left}} \biggl[ \left|P(\sigma_{\rm left} | \bm{x}) -  P(\sigma_{\rm right} | \bm{x})\right| \biggr]\\ 
    &\quad + \frac{1}{2} \mathbb{E}_{\bm{x} \sim P_{\rm right}}\biggl[ \left|P(\sigma_{\rm left} | \bm{x}) -  P(\sigma_{\rm right} | \bm{x})\right| \biggr] \\
    &= \frac{1}{2} \mathbb{E}_{\bm{x} \sim P_{\rm right}} \biggl[ \left(1+\frac{P_{\rm left}(\bm{x})}{P_{\rm right}(\bm{x})} \right)\left|P(\sigma_{\rm left} | \bm{x}) -  P(\sigma_{\rm right} | \bm{x})\right| \biggr]\\
    &= \frac{1}{2} \mathbb{E}_{\bm{x} \sim P_{\rm right}} \biggl[ P(\sigma_{\rm right} | \bm{x})\left(1+\frac{P_{\rm left}(\bm{x})}{P_{\rm right}(\bm{x})} \right)\left|1 - \frac{P(\sigma_{\rm left} | \bm{x})}{P(\sigma_{\rm right} | \bm{x})}\right| \biggr].\\
\end{align*}
Noting that $\frac{P(\sigma_{\rm left} | \bm{x})}{P(\sigma_{\rm right} | \bm{x})} = \frac{P_{\rm left}(\bm{x})}{P_{\rm right}(\bm{x})}$ and $P(\sigma_{\rm left} | \bm{x}) + P(\sigma_{\rm right} | \bm{x}) = 1$, we finally obtain
\begin{equation*}\label{eq:TV_proof_4}
    D_{g(x)= 1 - 2\min \{ x,1-x \} } = \mathbb{E}_{\bm{x} \sim P_{\rm right}} \biggl[ \frac{1}{2}\left|1 - \frac{P_{\rm left}(\bm{x})}{P_{\rm right}(\bm{x})}\right| \biggr] 
\end{equation*}
which corresponds to an $f$-divergence $D_{f}[P_{\rm left},P_{\rm right}]$
with $f(x) = \frac{1}{2}|1-x|$, i.e., the TV distance.

\paragraph{Freedom in Choice of $g$-function.} Note that the choice of $g$-function leading to a particular $g$-dissimilarity is not unique. In particular, we have that $D_{\tilde{g}} = D_{g}$ for any $\tilde{g}(x) = g(x) + c(\frac{1}{x} -2)$ where $c\in \mathbb{R}$ is some constant:
\begin{align*}
    D_{\tilde{g}} &= D_{g} + \frac{c}{2}\left(\mathbb{E}_{\bm{x} \sim P_{\rm left}} \biggl[ \frac{1-P(\sigma_{\rm left} | \bm{x})}{P(\sigma_{\rm left} | \bm{x})} - 1 \biggr] + \mathbb{E}_{\bm{x} \sim P_{\rm right}}\biggl[ \frac{1-P(\sigma_{\rm right} | \bm{x})}{P(\sigma_{\rm right} | \bm{x})} - 1
    \biggr] \right)\\
    &= D_{g} + \frac{c}{2}\left(\mathbb{E}_{\bm{x} \sim P_{\rm left}} \biggl[ \frac{P(\sigma_{\rm right} | \bm{x})}{P(\sigma_{\rm left} | \bm{x})} - 1 \biggr] + \mathbb{E}_{\bm{x} \sim P_{\rm right}}\biggl[ \frac{P(\sigma_{\rm left} | \bm{x})}{P(\sigma_{\rm right} | \bm{x})} - 1
    \biggr] \right)\\ 
    &= D_{g} + \frac{c}{2}\left( \mathbb{E}_{\bm{x} \sim P_{\rm left}} [ 1 ] + \mathbb{E}_{\bm{x} \sim P_{\rm right}} [ 1 ] -2\right) = D_{g}.
\end{align*}

\paragraph{Relation to the Fisher information.} In the following, we prove that any $g$-dissimilarity with $g(\frac{1}{2})=0$ and a twice-differentiable $g$-function reduces to the Fisher information in lowest order.

For this, consider the case with $L=1$ where we compare the distributions at two points in parameter space that are separated by $\delta T$. The corresponding $g$-dissimilarity is equivalent to an $f$-divergence $D_{f}[P(\bm{x}|T),P(\bm{x}|T+\delta T)]$. Note that $D_{f}[P(\bm{x}|T),P(\bm{x}|T)] = 0$ if $f(1)=0$ and $\left. \partial D_{f}[P(\bm{x}|T),P(\bm{x}|B)] /\partial B \right|_{B=T} = f'(1) \partial  \mathbb{E}_{\bm{x} \sim P(\cdot|T)} [ 1 ]/\partial T = f'(1) \partial 1/\partial T = 0$. The second-order derivative corresponds to
\begin{align*}
  \left. \frac{\partial^2 D_{f}[P(\bm{x}|T),P(\bm{x}|B)]} {\partial B^2 } \right|_{B=T} &= f'(1) \mathbb{E}_{\bm{x} \sim P(\cdot|T)} \biggl[  \frac{1}{P(\bm{x}|T)}\frac{\partial^2 P(\bm{x}|T)}{\partial T^2}
  \biggr]\\ 
  &\quad + f''(1) \mathbb{E}_{\bm{x} \sim P(\cdot|T)} \biggl[ \left(\frac{\partial \log P(\bm{x}|T)}{\partial T}\right)^2
  \biggr] \\
  &= f''(1) \mathcal{F}(T)
\end{align*}
assuming $f'(1)=0$, where $\mathcal{F}$ is the Fisher information. Thus, for $L=1$ and parameter values separated by $\delta T$, we can express any $g$-dissimilarity with $g(\frac{1}{2})=0$ as $D_{g} = \frac{g''(\frac{1}{2})}{32}\mathcal{F}(T)\delta T^2 + \mathcal{O}(\delta T^3)$. The fact that $f'(1)$ must be zero translates into the condition that $g'(\frac{1}{2})=0$. This is not a fundamental restriction since we have some freedom in the choice of $g$ function. That is, we can replace $g \mapsto \tilde{g}$, where $\tilde{g}(x) = g(x) + c(\frac{1}{x}-2)$ with $c=\frac{1}{6}g'(\frac{1}{2})$, retaining $D_{g} = D_{\tilde{g}}$ and ensuring that $\tilde{g}'(\frac{1}{2})=0$. 

\setcounter{equation}{0}
\setcounter{figure}{0}
\setcounter{table}{0}
\makeatletter
\renewcommand{\thesection}{C}
\renewcommand{\theequation}{C\arabic{equation}}
\renewcommand{\thefigure}{C\arabic{figure}}
\section{Details on Energy-Based Analysis of Temperature Transition}\label{app:energy_based}

Let $\bm{x} = (x_1, \dots, x_N)$ be a sequence of $N$ tokens generated for a fixed prompt from an autoregressive LLM such as the ones considered in this article. The distribution of $\bm{x}$ is given by
\begin{equation*}\label{eq:overall_prob}
    P(\bm{x}|T) = Q_{T}(x_N|x_1, \dots, x_{N-1})Q_{T}(x_{N-1}|x_1, \dots, x_{N-2}) \cdots Q_{T}(x_2|x_1) Q_{T}(x_1),
\end{equation*}
denoting the fact that the tokens are sampled sequentially. In each step, a token is sampled from a Boltzmann distribution $Q_{T}$, 
\begin{equation*}
    Q_{T}(x_{i}|x_1, \dots, x_{i-1}) = e^{- E(x_{i}|x_1, \dots, x_{i-1})/T}/Z_{i}(T). 
\end{equation*}
Here, $Z_{i}(T) = \sum_{x_{i}} e^{- E(x_{i}|x_1, \dots, x_{i-1})/T}$ is a normalization factor with the sum running over all possible $i$th tokens. The conditional energies, $E(x_{i}|x_1, \dots, x_{i-1})$ are typically referred to as logits and learned from data. Note that while the distribution over individual tokens can be expressed as a Boltzmann distribution at varying temperature, the overall distribution $P(\bm{x}|T)$ cannot.\footnote{In order for a quantity to be a valid energy of a system, it cannot itself depend on temperature, i.e., change with temperature.}

Nevertheless, we can define an energy scale for the entire system by viewing the overall probability distribution at $T=1$ as a Boltzmann
\begin{equation}\label{eq:E_tot_def1}
    P(\bm{x}|T=1) = e^{-E(\bm{x})}/Z,
\end{equation}
where $Z$ is a normalization constant independent of $\bm{x}$ (also referred to as partition function). Recall that any valid probability distribution can be written in the form of Eq.~\eqref{eq:E_tot_def1} with a suitably chosen energy function. Taking the logarithm of Eq.~\eqref{eq:E_tot_def2} and reordering, we have
\begin{equation}\label{eq:E_tot_def2}
    E(\bm{x}) =  - \log P(\bm{x}|T=1)  - \log Z.
\end{equation}
Using Eq.~\eqref{eq:E_tot_def2}, we can compute the total energy up to the constant $- \log Z$ which serves as our reference point for the energy scale. In the main text, we use $- \log P(\bm{x}|T=1)$ as the total energy to compute and compare energy statistics at various temperatures. In particular, the heat capacity remains the same under the mapping $- \log P(\bm{x}|T=1)  - \log Z \mapsto - \log P(\bm{x}|T=1)$.

\newpage
\bibliographystyleSI{unsrt}
\bibliographySI{refs_appendix}

\end{document}